\documentclass[
]{ceurart}

\usepackage{listings}
\usepackage{booktabs}
\usepackage{arydshln}
\usepackage{xcolor}
\begin{document}

\conference{}
%The 25th International Conference on Knowledge Engineering and Knowledge Management (EKAW 2026) - Satellite Events (Workshops, Tutorials, Demos, Posters). September 29 -- October 1, 2026. Turin, Italy.}

\title{ROBE: Reversed-Order-Biased-Experts for Extracting Extreme Long-tail Events from Historical Texts}

\author[1,2]{Stella Verkijk}[
orcid=0009-0000-9263-2272,
email=s.verkijk@vu.nl
]
\cormark[1]
\address[1]{Vrije Universiteit Amsterdam}
\address[2]{Huygens Instituut}

\author[1]{Piek Vossen}[
orcid=0000-0002-6238-5941,
email=p.t.j.m.vossen@vu.nl
]

\cortext[1]{Corresponding author.}

\begin{abstract}
  This paper proposes methods to extract over 50 types of events from a Dutch historical corpus spanning the 17th and 18th centuries. The methods we propose aim to tackle a very challenging scenario in Machine Learning: extracting the long-tail of the long-tail. Historic data from before the 19th century is in itself a niche domain not covered in the pre-training of Large Language Models, and we aim to extract events only scarcely annotated in the training data available for this domain. We propose creating expert classifiers for subgroups of the events present in the training data. We make these groupings based on similar frequency in the training data or on semantic relatedness. Experts trained on underrepresented events are assigned higher priority when predicting to avoid being dominated by frequency biases. We refer to this new way of combining classifiers, specifically tailored to protect the long-tail, as ROBE: Reversed-Order-Biased-Experts. We also propose a controlled method to create domain-specific synthetic data.\ Our two implementations of ROBE outperform a simple fine-tuned encoder model with a .16 increase in precision and a .05 increase in recall respectively. The best model achieves a .11 increase in f1 for a group of long-tail classes in our niche data set. 
\end{abstract}


\maketitle

\section{Introduction}
Even in the age of huge models, extracting long-tail information from free text remains a massive problem. The term \textquote{long-tail} can be interpreted in different ways. Any information from a niche domain can be seen as long-tail because it is infrequent in comparison to information widely available on the web. Similarly, any low-resource language can be considered long-tail. Additionally, the less frequent entities (those persons, places, events, etc. that occur sporadically) in a data set are considered to be long-tail. In this study, we propose a solution to deal with the long-tail of the long-tail, as we extract events from the archives of the Dutch East India Company (VOC) (1602-1799), handwritten in Early Modern Dutch, that show a very skewed distribution in the training data set. 

We formulate this task as a multiclass token classification problem (see Figure 1). Our dataset features 70 label types with the most frequent label occurring 794 times and the least frequent label once. The main problems we face are the skewed data distribution and the relatively large number of scarce data classes. We therefore build expert models for groups of event classes and apply these in a reversed order of bias: a Reversed Order of Biased Experts (ROBE) approach. We split the dataset into groups of labels that either have a similar frequency distribution in the training data or are semantically similar. We fine-tune a model for each group of labels, which we refer to as an expert. Each expert is also assigned a priority. Priority is assigned linearly where the expert trained on the labels with the smallest representation in the training data gets the highest priority and the expert trained on the labels with the biggest representation gets lowest priority. We then let each expert label our test dataset and resolve disagreements between event classes by choosing the label predicted by the model with the highest priority. A complete motivation and description of our ROBE approach is given in Section \ref{sec:meth-me}.

Previous research has shown that neither monolingual Dutch encoders nor larger multilingual encoders are capable of representing the semantics of these VOC archives, concluding that they are unable to capture relevant generalizations even after fine-tuning on domain-specific data \cite{arnoult2025fine, verkijk2025language, verkijk2026out}. %Hence, general language models are unable to represent even the first two levels of long-tailness: a low-resource language in a niche domain. 
In view of this, our ROBE approach consists of different fine-tuned implementations of GloBERTise, an encoder model pre-trained solely on the archives of the VOC\footnote{\url{https://huggingface.co/globalise/GloBERTise}}.

\begin{figure}[ht!]
\centering
    \includegraphics[width=12.3cm, height=4.6cm]{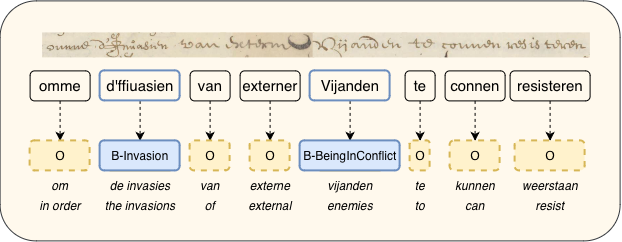}
    \caption{Example of event extraction task. In this example the nouns \textit{invasions} and \textit{enemies} (implicitly) refer to events, whereas the main verb \textit{resist} does not refer to any event from GLOBALISE's event tag set and is therefore not annotated. Recognizing that any part of speech can refer to an event means widening the variation space of tokens that can refer to a single event class, making some instances of events even more long-tail.}
    \label{fig:fig1}
\end{figure}

We show that compared to a model fine-tuned on all classes simultaneously without any data manipulation, a ROBE approach drastically improves precision, while a ROBE combined with a general fine-tuned model (ROBE+) improves both recall and precision. We also employ a specialized method to create synthetic data. Incorporating the resulting synthetic data in the fine-tuning process improves precision, but not recall.

This research gives insight into how to deal with a skewed dataset in a highly domain-specific and low-resource setting. We offer a solution that can be adopted with little to no extra resources or complicated engineering. We show that creating synthetic data can help if you have a way to infuse your data with domain-specific samples, but that splitting up your dataset and training models on specific label groups increases performance even more. We highlight that underperformance on long-tail knowledge can be mitigated without access to additional domain-specific resources or large generative models. We hope it inspires other researchers to devise new ways to tackle extreme long-tail information extraction.

\section{Related Work}
\textbf{Historical event extraction:}
There is some previous work related to event extraction in older data like ours \cite{borenstein2023multilingual, lai2021event, albuquerque2024applying}. Borrenstein et al. \cite{borenstein2023multilingual} aim to extract event information from historical newspaper adverts in different languages. However, they only extract attributes of events, and do not classify the events themselves. Lai et al. \cite{lai2021event} present BRAD, an annotated dataset of nineteenth century African American periodicals published from 1827 to 1909. Their task does involve event classification, but only of 12 event types. The authors report that transformer-based models applied on BRAD's event extraction task score substantially worse than when applied on the modern event dataset ACE \cite{walker2006ace}. Albuquerque et al. \cite{albuquerque2024applying} apply an event detection and classification system based on FrameNet for Portuguese \cite{sacramento2021joint} on 17th century letters, but do not report precision, recall or f1 scores. Other research that proposes automatic systems to extract events from historical texts was conducted on data from the 20th century \cite{sprugnoli2019novel, cybulska2011historical}.

A study investigating the performance of generative LLMs on event extraction reports that ChatGPT reaches only 51.04\% of the performance of a task-specific model in long-tail and complex scenarios \cite{gao2023exploring}. Kister \& Schirmer conclude that open-source LLMs can summarize Apartheid Witness Reports but lag behind when extracting entities from them (low precision being the main culprit), even though this dataset is in English and from 1996 \cite{kister-schirmer-2026-evaluating}. 

\textbf{Combining classifiers:}
Combining classifiers to solve a machine learning task is nothing new \cite{abimannan2023ensemble}. A popular method in Machine Learning is ensembling: training different classifiers on varying features of the dataset and combining the classifier's representations, for example by averaging over their embeddings, or more simply, combining their predictions with a majority vote scheme or a weighted average. ROBE does not combine all classifiers equally, but rather pushes classifiers representing the strongest biases prevalent in the training data to the background by overriding their predictions. Our approach is closer to what is known as a \textit{mixture of (competing) experts} \cite{aburomman2017survey, masoudnia2014mixture}. However, we do not use an extra classifier selection algorithm but rather use a simple rule-based approach to define the order of classifiers once. Also similar to our approach is \textit{bagging}, where classifiers are trained on subsets of the training data. In contrast to our approach, these subsets tend to be randomly sampled. 

\textbf{Synthetic data generation:}\label{sec:synth-relwork}
Creating synthetic data to enhance Machine Learning systems has been a common practice for years \cite{mikolajczyk2018data, maharana2022review}. By automatically creating data as training material, laborious human data annotation and curation can be avoided. Whereas earlier methods relied on, for example, Generative Adversarial Networks (GANs) \cite{dos2019data, akkem2024comprehensive}, the rise of high-performing generative Large Language Models has introduced a new era of synthetic data creation, especially for solving low-resource NLP problems \cite{meng2022generating, tan2024large, gao2022self, ye2022zerogen}. There are several advantages in using generative LLMs, such as the ease of prompting with natural language and, more importantly, the creativity of these probabilistic models, which makes it easier to spawn variety in the synthetic data. However, the inconsistency of LLMs, their largely unpredictable behavior, and their lack of knowledge of some specialized domains are downsides. Additionally, high frequency data and classes lead to generalizations adopted by a generative model that suppress long-tail interpretations. 

Although synthetic data augmentation can be seen as an established method, the speed at which the wide offer of LLMs that are now used for this method are being published, updated or adapted means the research into this method is also ongoing. Where some research focuses on evaluating different LLMs on their generation abilities \cite{kim2024evaluating}, or on how to actually evaluate the generated data \cite{ramesh2025synthtexteval}, and thereby the algorithms producing them \cite{havrilla2024surveying}, others focus on enhancing methods in prompting, for example by thinking of ways to generate data from existing datasets \cite{gandhi2024better}, using knowledge graphs \cite{kim2023soda} or entity samples \cite{yang2024synthetic}.

We draw on this last line of thinking by infusing our prompt templates with data points and (annotation) examples specific to our domain. By doing so, we ingrain domain-specific attributes into our synthetic data without the need to fine-tune or instruct a generative model. This allows us to steer the training towards long-tail phenomena. We describe this process in detail in Section \ref{sec:synth-method}.

\begin{figure}[h!]
    \includegraphics[trim={0cm 0cm 0cm 0.9cm}, clip, width=16cm, height=9cm]{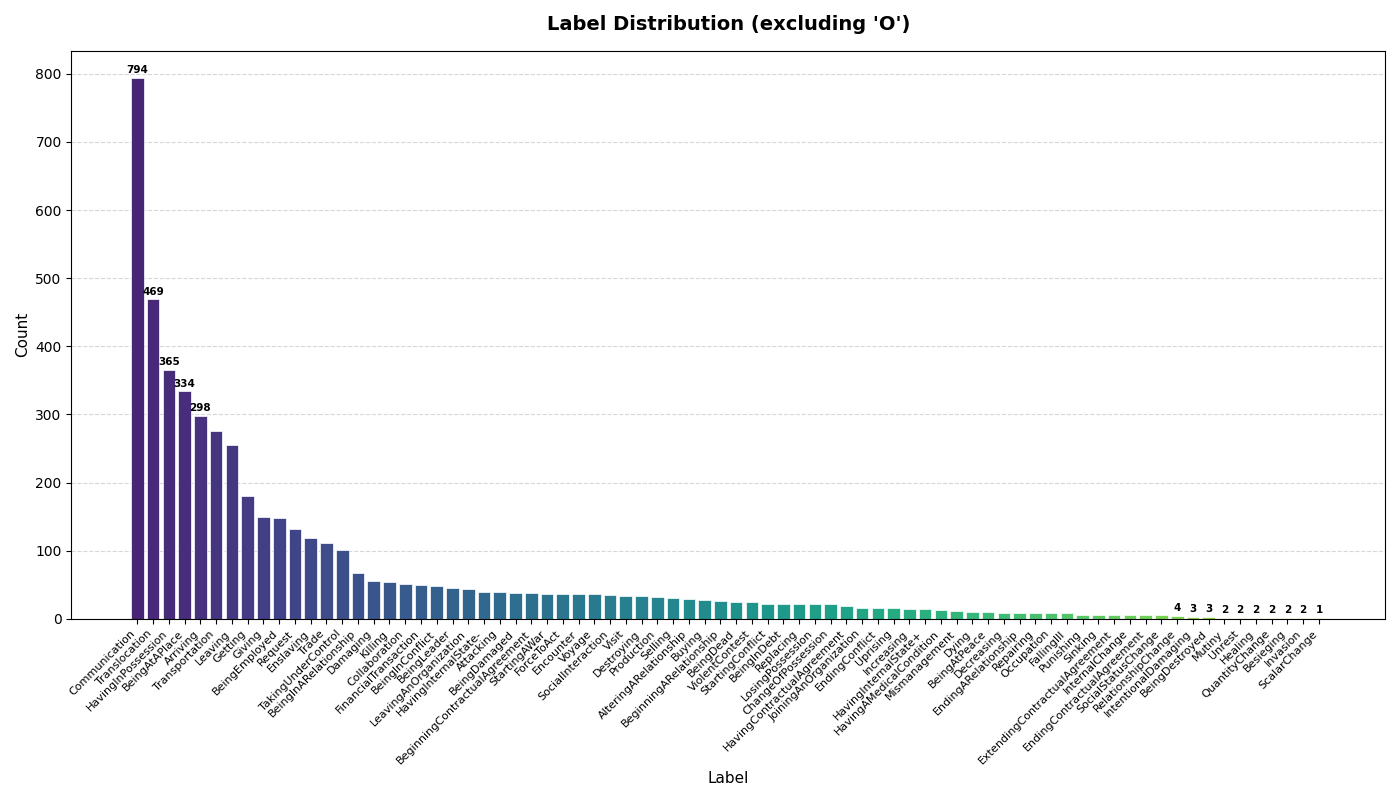}
    \caption{Label distribution in the expert annotated data of all event classes. Counts are per token.}
    \label{fig:labeldistribution2}
\end{figure}

\section{Data}

The data we work with was created and made publicly available by the GLOBALISE project. GLOBALISE uses their own event ontology as a backbone of their annotation process. This ontology organizes events taxonomically and makes relations between states (Static Events) and changes-of-state (Dynamic Events) explicit. We refer to the GLOBALISE papers on the ontology and the  annotation process for further details \cite{verkijk2023sunken, verkijk2024studying}. The training data consists of 280 annotated pages from 26 documents and the test data of 18 annotated pages from 3 documents. 

The task is different from classical event extraction tasks that rely on locating the predicate of the sentence: Any part of speech can refer to an event (see Figure \ref{fig:fig1}). The reasoning behind this task design is twofold: i) the archival texts contain very long sentences with unclear to no sentence boundaries, and ii) historically relevant information about events is often expressed implicitly, for example through nouns or adjectives, rather than only through verbs.

As an intuitive way to illustrate that events and entities that are central to the world of the VOC are low-resource in the current data climate, we can compare the number of hits for certain Google searches (measured on September 19, 2026). The Dutch word \textquoteleft{boot'} (\textit{boat}) returns 1,4 billion hits, while the word \textquoteleft{fregat'} (a common boat type used by the VOC) returns 1,5 million hits. The English word \textquoteleft{slavery'} returns 150 million hits and the word \textquoteleft{enslaving'} 2,3 million, while the word \textquoteleft{divorce'} returns 217 million hits and \textquoteleft{wedding'} 3 billion. Searching for \textquoteleft{Dutch East India Company'} returns 196 million hits, while \textquoteleft{Apple'} returns 6,7 billion hits. 

Zooming in on the data distribution within our domain, Figure \ref{fig:labeldistribution2} shows the highly skewed label distribution in the training data. The frequencies follow a Zipfian distribution. The difference between the most frequent event (\textit{Communication} with 794 tokens, 16\% of all tokens with an event label) and the least frequent (\textit{ScalarChange} with 1, 0.02\%) is extreme. Also, the difference between the most frequent event and the runner-up (\textit{Translocation}) is large: Translocation represents 9\% of all annotated events. 92\% of tokens do not refer to an event. 

The VOC archives are known to describe a lot of trade and politics, but also violence. The VOC was a colonial enterprise and political entity that spread war and slave labor. However, describing the violence was not the main focus of the employees of the VOC, as writing letters was more often motivated by a need for more money and resources from the homeland, the obligation to report on evolving relationships with local rulers, and other economically relevant topics. As we can see in Figure \ref{fig:labeldistribution2}, some of the events at the extreme end of the long-tail are \textit{Mutiny}, \textit{Besieging} and \textit{Invasion}. Other events that are typical of VOC violence, such as \textit{Enslaving}, \textit{TakingUnderControl}, \textit{Attacking} and \textit{ForceToAct} are also much less frequent than classes related to translocation and trade. Hence, we observe a cultural bias reflected in the data distribution. A system that is only able to extract frequent classes will inaccurately represent the history of the VOC and the ability to extract long-tail events is therefore essential.

\section{Method}

\subsection{The ROBE approach}\label{sec:meth-me}
The idea behind ROBE is to split the classification problem into smaller chunks (bootstrapping) by training different models on different groups of events. These groupings are based on their representation in the training data and on their semantics (putting semantically similar events in the same group). Grouping classes with similar representations avoids one class being pushed to the background by other classes with much more examples in the training data. 

A typical thing you see when training models on little data is that after a small amount of epochs, the model is only able to recognize the most common classes it has seen in the training data and starts predicting those any time it encounters something that refers to an event. The intuition is thus that grouping classes with similar representations in one task makes the model converge for each class at a similar moment during training. All models were fine-tuned with learning rate 5e-5. For further parameter settings we refer to our repository with all code and data to fine-tune models\footnote{\url{https://github.com/globalise-huygens/nlp-event-finetuning-experts}}.  

When training a model for a single group, all annotated labels of classes that were not included in the group are relabeled as \textquoteleft{O'}, similar to a one-against-all training regime. While this introduces the downside that each group has an even bigger \textquoteleft{O'} class, the impact of this is limited because the O class is already extremely dominant, with 92\% of tokens (see Figure \ref{fig:labeldistribution1} in the Appendix).

\begin{table}[!ht]
    \centering
    \small
    \begin{tabular}{p{7.8cm} c p{2cm} p{2cm} c c}
    \toprule
        Event classes in group & \makecell{\#classes} & \makecell{Grouped on\\repr. window} & \makecell{Grouped on\\semantics} & \makecell{\#e} & Prio \\
    \midrule
        Buying, BeginningARelationship, ViolentContest, BeingInDebt, Replacing, LosingPossession, ChangeOfPossession, JoiningAnOrganization, Unrest, HavingInternalState+, Mismanagement, HavingAMedicalCondition, BeingAtPeace, QuantityChange & 14 & 10--30 & n/a & 20 & \textbf{1} \\
        \addlinespace
        TakingUnderControl, Enslaving, Unrest, Attacking, StartingAWar, Killing, BeingInConflict, ViolentContest, ForceToAct & 9 & n/a & Violence, repression, conflict & 20 & \textbf{2} \\
        \addlinespace
        BeingInARelationship, Damaging, Killing, Collaboration, FinancialTransaction, BeingInConflict, BeingLeader, LeavingAnOrganization, HavingInternalState--, Attacking, BeingDamaged, HavingContractualAgreement, StartingAWar, ForceToAct, Encounter, SocialInteraction, Visit, BeingDestroyed, Production, Selling, BeingDead, RelationshipChange & 22 & 30--100 & n/a & 18 & \textbf{3} \\
        \addlinespace
        Getting, Giving, BeingEmployed, Request, Enslaving, Trade, TakingUnderControl & 7 & 100--200 & n/a & 12 & \textbf{4}\\
        \addlinespace
        HavingInPossession, Giving, Buying, Selling, ChangeOfPossession, Trade, FinancialTransaction, LosingPossession, Getting & 9 & n/a & Possession & 20 & \textbf{5} \\
        \addlinespace
        HavingInPossession & 1 & 300--400 & n/a & 20 & \textbf{6} \\
        \addlinespace
        Translocation, Transportation, Leaving, Arriving, Voyage, BeingAtAPlace & 6 & n/a & Translocation & 20 & \textbf{7} \\
        \addlinespace
        Communication & 1 & >700& n/a & 10 & \textbf{8} \\
    \bottomrule
    \end{tabular}
    \caption{Event classes grouped for training the experts of ROBE, given in order of priority. Groups with long-tail classes get higher priority. \#e = number of epochs the expert was trained; Prio = Priority. Grouped on repr. window = range of label frequency in the training data given in amount of tokens, not mentions.}
    \label{tab:event-classes}
\end{table}

See Table \ref{tab:event-classes} for an overview of all groups with their representation in the training data. For the ROBE experts, some classes are merged in order to increase their representation. Classes are only considered for merge if it concerns i) a related Dynamic and Static Event, ii) events that are subclasses of the same class in the taxonomy, or iii) events that are one step away from each other in the taxonomy. Examples of the first are \textit{Dying} and \textit{BeingDead}; \textit{Destroying} and \textit{BeingDestroyed}. Examples of the second are \textit{Besieging} and \textit{Invasion}, where both are subclasses of \textit{Attacking}. Examples of the third are \textit{EndingARelationship} and \textit{RelationshipChange}, where the first is a subtype of the second. Table \ref{tab:merged-classes} in the Appendix shows which classes are merged, what type of merge it is, and what their individual representation is. 

\subsection{Synthetic data creation}\label{sec:synth-method}
As explained in Section \ref{sec:synth-relwork}, previous studies on creating synthetic data have shown that infusing (domain-specific) knowledge from existing datasets can lead to more representative synthetic data for niche domains. 
Part of GLOBALISE's work involves creating structured data of persons, places, ships and commodities described in the archives (i.e., \cite{mourits2025modelling}). We use GLOBALISE's published dataset on the ``Namebooks" of the VOC to extract names of persons, their professions and the location they were mentioned \footnote{\url{https://dataverse.nl/dataset.xhtml?persistentId=doi:10.34894/MD59SC}}. Additionally, we use their published commodities thesaurus to extract names of commodities and their definitions\cite{pepping2024reflections}\footnote{\url{https://dataverse.nl/dataset.xhtml?persistentId=doi:10.34894/5PZPPK}}. Lastly, GLOBALISE provided a ship dataset\footnote{This data set is still under development and will be published in the future} from which we extract names of ship types and their descriptions. We then create prompt templates that combine these data like so:

\begin{quote}
\small
\textit{See the following example of an annotated sentence:} <example>\textit{. Create a sentence in Early Modern Dutch in which} <commodity> \textit{is transported. This is the definition of} <commodity>: <definition>\textit{. Write a sentence that could be found in the Archives of the Dutch East India Company (1602-1799). Make sure to adjust your style to the Dutch East India Company. Put the word that refers to the transportation in square brackets like in the example.}
\end{quote}

We create prompts by randomly sampling from the aforementioned datasets and a small selection of 9 example annotations to create several combinations for each prompt template. Some prompt templates do not use an example annotation to promote diversity in the output. We use a total of 8 prompt templates. Table \ref{tab:prompts} in the Appendix contains all the prompt templates used and Table \ref{tab:examples} all annotation examples. Using the created prompt templates between 15 to 30 times we create 375 annotated synthetic sentences for \textit{Translocation} and \textit{Transportation} with gpt-4o. We chose \textit{Translocation} and \textit{Transportation} because these are the most appropriate events to infuse with domain-specific entities like locations, ships and commodities. It serves as a proof-of-concept: If results show incorporating this type of synthetic data positively impacts performance on these classes, we can consider doing something similar for more extreme long-tail classes in the future\footnote{All code and the resulting synthetic data can be found at \url{https://github.com/globalise-huygens/nlp-event-synthetic-data-generation}.}.

\subsection{Overview of evaluated methods}
Table \ref{tab:modeloverview} gives an overview of all methods evaluated against the test set. The End to End model is simple: GloBERTise fine-tuned once on all available human-annotated training data without any data manipulation (EtE). We also fine-tune a version of this model on the training data augmented with our synthetic data. Apart from ROBE as described in Section \ref{sec:meth-me}, we also propose a version of ROBE where we include EtE as an expert: ROBE+. EtE gets lowest priority in ROBE+ (i.e., place 9).

The lexicon and the GenLLM serve as baselines: the first representing the extreme end of domain-specificity and human labor and the second representing the extreme end of general pre-trained models that do not require any human annotations. As per the previous research mentioned so far indicating general pre-trained models do not perform well on long-tail domains \cite{verkijk2025language, arnoult2025fine, gao2022self, kister-schirmer-2026-evaluating} as well as research that shows that Generative LLMs do not fare well when applied to classification problems where the label set is large \cite{khrapunova2025bridging}, we do not expect the GenLLM to perform well. From the lexicon we can expect more, as it was created in an iterative manner, using seeds from human annotations and expanding with a specialized Word2Vec model. Furthermore, contributions to the lexicon were given by annotators who also annotated the test set, so the possibility that the lexicon is overfit on the test set should be taken into account. Annotators added words to the lexicon over a time span of years\footnote{See GLOBALISE's blog post on their lexical approach:  \url{https://globalise.huygens.knaw.nl/yes-okay-but-what-were-they-doing/}. We used the second release of the lexicon, which is available at \url{https://github.com/globalise-huygens/nlp-event-lexical-approach/blob/main/lexicon_release2.csv.}}.

\begin{table}[!ht]
    \centering
    \small
    \begin{tabular}{lll}
    \toprule
        Approach & Acronym & Design \\ \hline
        Lexicon & Lexicon & Lexical Baseline provided by GLOBALISE\\
        gpt-5.1 & GenLLM & Generative baseline using gpt-5.1, zero-shot and few-shot setting \\
        End to End & EtE & Fine-tuned GloBERTise on all data  \\ 
        End to End with augmented data & EtE+synth & EtE trained on all data plus additional synthetic data \\
        Reversed-Order-Biased-Expert & ROBE & Eight fine-tuned GloBERTises on different label groups  \\ 
        ROBE with EtE & ROBE+ & Nine fine-tuned GloBERTises, of which one EtE \\ 
        ROBE with EtE and lexicon & ROBE+Lex& Nine fine-tuned GloBERTises, of which one EtE, and a lexicon \\ 
       \bottomrule
    \end{tabular}
    \caption{Overview of evaluated methods}\label{tab:modeloverview}
\end{table}

\subsection{Evaluation}
GLOBALISE released three documents annotated by at least four annotators as test data. We adjudicated these annotations by extracting the majority vote where possible, and otherwise consulting the annotation guidelines and consulting historians. We kept track of our progress and concluded that the percentage of adjudications that could be done through a majority vote was only 47\%. Of those, the percentage where the majority vote (three out of four annotations) was \textquoteleft{O'} was 52\%. Additionally, the percentage of manual adjudications that could be made on ontological basis was only 12\%. Hence, when annotators disagreed between event types, there was often no simple way to adjudicate them. These statistics show that the annotation is partially subjective, a phenomenon that is encountered often in NLP \cite{uma2021learning}. We can expect that evaluating strictly on the adjudicated labels may result in relatively low scores. 

%We therefore evaluate both on the adjudicated and the non-adjudicated test set. For the latter, we count a prediction of the system as correct if it is the same as one of the gold labels. 

\section{Results}\label{sec:results}

\subsection{Evaluation on test set}

\begin{table}[ht]
\centering
\resizebox{\textwidth}{!}{%
\begin{tabular}{lccc:ccc:ccc:ccc:ccc:c}
\toprule
 & \multicolumn{3}{c}{Lexicon} & \multicolumn{3}{c}{EtE} & \multicolumn{3}{c}{EtE+synth} & \multicolumn{3}{c}{ROBE} & \multicolumn{3}{c}{ROBE+} & \\
\cmidrule(lr){2-4} \cmidrule(lr){5-7} \cmidrule(lr){8-10} \cmidrule(lr){11-13} \cmidrule(lr){14-16}
 & P & R & F1 & P & R & F1 & P & R & F1 & P & R & F1 & P & R & F1 & Supp. \\
\midrule
overall & \textbf{0.69} & 0.14 & 0.23 & 0.29 & \underline{0.20} & 0.24 & 0.42 & 0.19 & 0.26 & \underline{0.45} & \underline{0.20} & \textbf{0.28} & 0.31 & \textbf{0.25} & 0.27 & 626 \\
translocation & \textbf{0.77} & 0.25 & 0.38 & 0.66 & 0.59 & 0.62 & \underline{0.75} & 0.49 & 0.59 & 0.74 & 0.51 & 0.60 & 0.66 & \textbf{0.65} & \textbf{0.65} & 208 \\
violence & \textbf{1.00} & 0.03 & 0.05 & 0.20 & 0.04 & 0.07 & \underline{0.40} & \underline{0.08} & \underline{0.14} & 0.38 & \underline{0.08} & 0.13 & 0.31 & \textbf{0.12} & \textbf{0.18} & 73 \\
\bottomrule
\end{tabular}%
}
\caption{Micro average results on the adjudicated test data across event classes for EtE, EtE+synth, ROBE, ROBE+, and the lexical baseline. For the overall scores, we apply strict evaluation on BIO labels and event types (a prediction is only correct if it has the correct BIO label and the correct event type). For both the scores on translocation and violence, we count a prediction as correct if it falls within the set of labels that refer to either translocation or violence (as specified in Table \ref{tab:event-classes}. BIO labels do not have to overlap. To exemplify: if the gold label is \textit{B-Translocation} and the model predicts \textit{I-Transportation}, we count it as correct. We do not include the \textquoteleft{O'} class in any metric.}
\label{tab:results}
\end{table}

As we can see in Table \ref{tab:results}, the lexicon offers a strong baseline, scoring highest in precision across the board. ROBE scores .16 higher in precision over EtE overall and .18 for the more challenging group of labels referring to violence, repression and conflict. Recall also increases slightly for this group. Recall does not increase in the overall score and decreases for the translocation labels. Our combined model, ROBE+, scores highest in recall and f1 in most evaluation settings. The EtE model trained with additional synthetic data drastically improves precision over EtE, but declines in recall. We think this might be caused by the fact that the synthetic data contains unlabelled mentions of other events than \textit{Transportation} and \textit{Translocation}, providing the model with negative examples for classes it should recognize. It might also be that the generated output is limited in its variety in linguistic realizations of events. The EtE model trained on augmented data is outperformed in its f1 score for translocation classes by both ROBE and ROBE+. Training an expert translocation model on the augmented dataset, and replacing the original translocation expert with the synthetic expert in the ROBE model does not increase scores for ROBE or ROBE+, as can be seen in Table \ref{tab:me_combined_synth} in the Appendix\footnote{We make all trained experts and the EtE model publicly available at \url{https://huggingface.co/collections/StellaVerkijk/robe-event-classifiers-for-globalise} and both models trained on synthetic data at \url{https://huggingface.co/collections/StellaVerkijk/event-classifiers-trained-on-synthetically-augmented-data}}. Gpt5.1 performs poorly overall and for the isolated translocation labels. It scores on the higher end for violence classes, but not very reliably, as f1 scores differ while the temperature was set to 0, as shown in Table \ref{tab:gpt-results} in the Appendix.

Unfortunately, the classes in the lowest frequency group, which we trained the classifier with the highest priority on, do not have sufficient support in the test data to report reliable results on. The only class in this group with a support of more than 5 tokens in the adjudicated data is \textit{HavingInternalState+} with a support of 11. The lexicon and EtE models have an f1 of 0 for this class across the board, while both ROBE and ROBE+ have an f1 of 0.15, but this score is only based on the same 2 predictions by each model.  

To give our best performing model a little boost, we can include our lexicon as the lowest-priority classifier: the lexicon is used to label events missed by any of the classifiers in ROBE+ (i.e., ROBE+ labels a string as \textquoteleft{O'} but the lexicon does not). This improves scores slightly. 

\begin{table}[ht]
\centering
\begin{tabular}{lccc:c}
\toprule
& \multicolumn{2}{c}{ROBE+ with Lexicon} & & \\
\cmidrule(lr){1-5}
 & P & R & F1 & Supp. \\
\midrule
overall & 0.32 & \textbf{0.27} & \textbf{0.29} & 626 \\
translocation & 0.65 & \textbf{0.69} & \textbf{0.67} & 208 \\
violence & 0.31 & 0.12 & 0.18 & 73 \\
\bottomrule
\end{tabular}%
\caption{Micro average results on the adjudicated test set across event classes for ROBE+ combined with the lexicon. Scores are in boldface if they surpass any of the scores presented in Table\ref{tab:results}. Metrics calculated as in Table \ref{tab:results}.}\label{tab:robe+lex}
\end{table}

The results indicate that assigning higher priority to classifiers that are experts on long-tail events does not hurt performance on events that are not long-tail. Experts on subsets of the dataset are more conservative and improve precision. Combining experts with a general model trained on the complete dataset balances out precision and recall\footnote{Code and data to reproduce results can be found at \url{https://github.com/globalise-huygens/nlp-event-classifier-evaluation/tree/main}}.

\subsection{External Validation}\label{sec:external}
We also gathered external user validation of a simplified version of the ROBE model\footnote{A ROBE without experts for possession and violence (i.e., missing models 2 and 5) and with a slight difference in priorities, namely 7-6-1-3-4, favoring the translocation model and the one built on the 300-400 token frequency window.}. These external validation data were gathered during two iterations of a workshop for historians about Machine Learning. Participants were asked to assign one of two values to each prediction of ROBE: \textit{useful} or \textit{misleading}. See the results of the simplified ROBE on the part of the test set that validators judged in Table \ref{tab:old_me}. 

In the first iteration of the workshop, the model’s predictions were labeled as useful \textbf{75\%} of the time (average of six validators). This is based on a total of 329 validated (non unique) predictions. Of the validators, three had 1-5 years of experience reading archives, two had one year of experience and one had none. One validator was discarded because they did not pass a translation task made to assess their level of reading Early Modern Dutch. Interestingly, the participants labeled our adjudicated gold data as useful 68\% of the time (based on a total of 78 validated gold labels). This reiterates that we are dealing with a complicated task where multiple interpretations of the text are possible. Three out of six annotators did not validate any gold data (they did not reach that part of the assignment).

\begin{table}[h]
\begin{tabular}{lcccc}
\toprule
 & P & R & F1 & Support \\
\midrule
overall & 0.38 & 0.13 & 0.20 & 260 \\
translocation & 0.75 & 0.55 & 0.64 & 78 \\
violence & 0.00 & 0.00 & 0.00 & 19 \\
\bottomrule
\end{tabular}
\caption{Micro average results on the adjudicated test set for a simplified ROBE on part of the test set. Metrics calculated as in Table  \ref{tab:results}.}
\label{tab:old_me}
\end{table}

In the second iteration (9 participants), ROBE's predictions were labeled useful \textbf{79\%} of the time, based on a total of 578 validated predictions. In this group, there was one person with 1-5 years of experience reading archives, one with less than half a year, and the rest had none. They labeled our adjudicated gold data as useful 84\% of the time, which was based on a total of 190 validated gold labels. One person did not validate any gold data\footnote{For the data see \url{https://github.com/globalise-huygens/nlp-event-external-machine-output-validation} and for workshop slides see \url{https://zenodo.org/records/20118595}}. 

Since validators only judged events predicted by ROBE, they did not have any way to judge events ROBE might have missed, hence, we are only validating the model's precision, and we cannot say anything about the recall. However, the conclusion we can draw is that in the external validation, ROBE scores much higher on precision than in our classical evaluation, as can be seen in Table \ref{tab:old_me}. We believe this finding contextualizes the somewhat low scores across the board in Table \ref{tab:results}. %It is plausible we are dealing with possible false false positives. 

\section{Conclusion}\label{sec:discuss}
Our paper offers an accessible approach to enhance performance when classifying long-tail events in a low-resource language and niche domain. By combining classifiers and assigning a fixed picking order, where classifiers trained on underrepresented labels in the training data receive highest priority, it is possible to improve performance across classes in a skewed dataset and balance out precision and recall. 

Our paper has a few limitations. An analysis of the synthetic data would give more insight into why it improves precision but not recall. Also, a more fine-grained error analysis where we compare predictions of different models on tokens that refer to specific long-tail events in the test set would provide a clearer analysis of the difference between model configurations.

\section{Declaration on Generative AI use}
Generative AI was sometimes consulted in the process of writing some of the software presented in this study. If a complete function was written by AI, we state this in the code's documentation. We did not use Generative AI for any of the categories described in CEUR's GenAI Usage Taxonomy.

\begin{acknowledgments}
This research was funded by the Dutch Research Council (NWO) under project number
175.2019.003 and is part of the Globalise
project. All fine-tuning took place
on the National Supercomputer Snellius, which
we were granted access to by the Dutch national
e-infrastructure with the support of the SURF Cooperative using grant no. EINF-11854. We want to thank Kay Pepping for his help conceptualising and setting up the synthetic data creation pipeline and Lodewijk Petram for proofreading.
\end{acknowledgments}

\bibliography{bib}

@article{meng2022generating,
  title={Generating training data with language models: Towards zero-shot language understanding},
  author={Meng, Yu and Huang, Jiaxin and Zhang, Yu and Han, Jiawei},
  journal={Advances in Neural Information Processing Systems},
  volume={35},
  pages={462--477},
  year={2022}
}

@article{dos2019data,
  title={Data Augmentation Using GANs},
  author={dos Santos Tanaka, Fabio Henrique Kiyoiti and Aranha, Claus},
  journal={Proceedings of Machine Learning Research XXX},
  volume={1},
  pages={16},
  year={2019}
}

@inproceedings{mikolajczyk2018data,
  title={Data augmentation for improving deep learning in image classification problem},
  author={Miko{\l}ajczyk, Agnieszka and Grochowski, Micha{\l}},
  booktitle={2018 international interdisciplinary PhD workshop (IIPhDW)},
  pages={117--122},
  year={2018},
  organization={IEEE}
}

@article{maharana2022review,
  title={A review: Data pre-processing and data augmentation techniques},
  author={Maharana, Kiran and Mondal, Surajit and Nemade, Bhushankumar},
  journal={Global Transitions Proceedings},
  volume={3},
  number={1},
  pages={91--99},
  year={2022},
  publisher={Elsevier}
}

@article{akkem2024comprehensive,
  title={A comprehensive review of synthetic data generation in smart farming by using variational autoencoder and generative adversarial network},
  author={Akkem, Yaganteeswarudu and Biswas, Saroj Kumar and Varanasi, Aruna},
  journal={Engineering Applications of Artificial Intelligence},
  volume={131},
  pages={107881},
  year={2024},
  publisher={Elsevier}
}

@article{tan2024large,
  title={Large language models for data annotation and synthesis: A survey},
  author={Tan, Zhen and Li, Dawei and Wang, Song and Beigi, Alimohammad and Jiang, Bohan and Bhattacharjee, Amrita and Karami, Mansooreh and Li, Jundong and Cheng, Lu and Liu, Huan},
  journal={arXiv preprint arXiv:2402.13446},
  year={2024}
}

@article{gao2022self,
  title={Self-guided noise-free data generation for efficient zero-shot learning},
  author={Gao, Jiahui and Pi, Renjie and Lin, Yong and Xu, Hang and Ye, Jiacheng and Wu, Zhiyong and Zhang, Weizhong and Liang, Xiaodan and Li, Zhenguo and Kong, Lingpeng},
  journal={arXiv preprint arXiv:2205.12679},
  year={2022}
}

@inproceedings{ye2022zerogen,
  title={Zerogen: Efficient zero-shot learning via dataset generation},
  author={Ye, Jiacheng and Gao, Jiahui and Li, Qintong and Xu, Hang and Feng, Jiangtao and Wu, Zhiyong and Yu, Tao and Kong, Lingpeng},
  booktitle={Proceedings of the 2022 Conference on Empirical Methods in Natural Language Processing},
  pages={11653--11669},
  year={2022}
}

@article{kim2024evaluating,
  title={Evaluating Language Models as Synthetic Data Generators},
  author={Kim, Seungone and Suk, Juyoung and Yue, Xiang and Viswanathan, Vijay and Lee, Seongyun and Wang, Yizhong and Gashteovski, Kiril and Lawrence, Carolin and Welleck, Sean and Neubig, Graham},
  journal={CoRR},
  year={2024}
}

@article{ramesh2025synthtexteval,
  title={SynthTextEval: Synthetic Text Data Generation and Evaluation for High-Stakes Domains},
  author={Ramesh, Krithika and Smolyak, Daniel and Zhao, Zihao and Gandhi, Nupoor and Agarwal, Ritu and Bjarnad{\'o}ttir, Margr{\'e}t and Field, Anjalie},
  journal={arXiv preprint arXiv:2507.07229},
  year={2025}
}

@article{havrilla2024surveying,
  title={Surveying the effects of quality, diversity, and complexity in synthetic data from large language models},
  author={Havrilla, Alex and Dai, Andrew and O'Mahony, Laura and Oostermeijer, Koen and Zisler, Vera and Albalak, Alon and Milo, Fabrizio and Raparthy, Sharath Chandra and Gandhi, Kanishk and Abbasi, Baber and others},
  journal={arXiv preprint arXiv:2412.02980},
  year={2024}
}

@inproceedings{gandhi2024better,
  title={Better synthetic data by retrieving and transforming existing datasets},
  author={Gandhi, Saumya and Gala, Ritu and Viswanathan, Vijay and Wu, Tongshuang and Neubig, Graham},
  booktitle={Findings of the Association for Computational Linguistics: ACL 2024},
  pages={6453--6466},
  year={2024}
}

@inproceedings{kim2023soda,
  title={Soda: Million-scale dialogue distillation with social commonsense contextualization},
  author={Kim, Hyunwoo and Hessel, Jack and Jiang, Liwei and West, Peter and Lu, Ximing and Yu, Youngjae and Zhou, Pei and Bras, Ronan and Alikhani, Malihe and Kim, Gunhee and others},
  booktitle={Proceedings of the 2023 Conference on Empirical Methods in Natural Language Processing},
  pages={12930--12949},
  year={2023}
}

@article{yang2024synthetic,
  title={Synthetic continued pretraining},
  author={Yang, Zitong and Band, Neil and Li, Shuangping and Candes, Emmanuel and Hashimoto, Tatsunori},
  journal={arXiv preprint arXiv:2409.07431},
  year={2024}
}

@article{khrapunova2025bridging,
  title={Bridging Zero-Shot and Fine-Tuned Performance in Text Classification throughRetrieval-Augmented Prompting},
  author={Khrapunova, Olesia},
  journal={American Academic Scientific Research Journal for Engineering, Technology, and Sciences},
  year={2025}
}

@inproceedings{mourits2025modelling,
  title={Modelling the enslaved as historical persons: Extending the Persons in Context (PiCo) model to fit enslaved individuals},
  author={Mourits, Rick and Pepping, Kay and van Oort, Thunnis and Konings, Pascal and van Duijvenvoorde, Britt},
  booktitle={Digital Humanities Benelux 2025},
  year={2025}
}

@article{uma2021learning,
  title={Learning from disagreement: A survey},
  author={Uma, Alexandra N and Fornaciari, Tommaso and Hovy, Dirk and Paun, Silviu and Plank, Barbara and Poesio, Massimo},
  journal={Journal of Artificial Intelligence Research},
  volume={72},
  pages={1385--1470},
  year={2021}
}

@inproceedings{verkijk2025language,
  title={Language Models Lack Temporal Generalization and Bigger is Not Better},
  author={Verkijk, Stella and Vossen, Piek and Sommerauer, Pia},
  booktitle={Findings of the Association for Computational Linguistics: ACL 2025},
  pages={20629--20637},
  year={2025}
}

@article{arnoult2025fine,
  title={Fine-grained Named-Entity Recognition for the East-India Company domain},
  author={Arnoult, Sophie and Nijman, Brecht and Wissen, Leon van},
  journal={Anthology of Computers and the Humanities},
  volume={3},
  pages={953--967},
  year={2025}
}

@inproceedings{verkijk2026out,
  title={Out-of-Tune rather than Fine-Tuned: How Pre-training, Fine-tuning and Tokenization Affect Semantic Similarity in a Historical, Non-Standardized Domain},
  author={Verkijk, Stella and Vossen, Piek},
  booktitle={Proceedings of the Second Workshop on Language Models for Low-Resource Languages (LoResLM 2026)},
  pages={515--531},
  year={2026}
}

@inproceedings{verkijk2023sunken,
  title={Sunken Ships Shan't Sail: Ontology Design for Reconstructing Events in the Dutch East India Company Archives.},
  author={Verkijk, Stella and Vossen, Piek},
  booktitle={CHR},
  pages={320--332},
  year={2023}
}

@inproceedings{verkijk2024studying,
  title={Studying Language Variation Considering the Re-Usability of Modern Theories, Tools and Resources for Annotating Explicit and Implicit Events in Centuries Old Text},
  author={Verkijk, Stella and Sommerauer, Pia and Vossen, Piek},
  booktitle={Proceedings of the Eleventh Workshop on NLP for Similar Languages, Varieties, and Dialects (VarDial 2024)},
  pages={174--187},
  year={2024}
}

@article{sprugnoli2019novel,
  title={Novel event detection and classification for historical texts},
  author={Sprugnoli, Rachele and Tonelli, Sara},
  journal={Computational Linguistics},
  volume={45},
  number={2},
  pages={229--265},
  year={2019}
}

@inproceedings{cybulska2011historical,
  title={Historical event extraction from text},
  author={Cybulska, Agata and Vossen, Piek},
  booktitle={Proceedings of the 5th ACL-HLT Workshop on Language Technology for Cultural Heritage, Social Sciences, and Humanities},
  pages={39--43},
  year={2011}
}

@inproceedings{borenstein2023multilingual,
  title={Multilingual event extraction from historical newspaper adverts},
  author={Borenstein, Nadav and da Silva Perez, Nat{\'a}lia and Augenstein, Isabelle},
  booktitle={Proceedings of the 61st Annual Meeting of the Association for Computational Linguistics (Volume 1: Long Papers)},
  pages={10304--10325},
  year={2023}
}

@inproceedings{lai2021event,
  title={Event extraction from historical texts: A new dataset for black rebellions},
  author={Lai, Viet Dac and Van Nguyen, Minh and Kaufman, Heidi and Nguyen, Thien Huu},
  booktitle={Findings of the Association for Computational Linguistics: ACL-IJCNLP 2021},
  pages={2390--2400},
  year={2021}
}

@article{gao2023exploring,
  title={Exploring the feasibility of chatgpt for event extraction},
  author={Gao, Jun and Zhao, Huan and Yu, Changlong and Xu, Ruifeng},
  journal={arXiv preprint arXiv:2303.03836},
  year={2023}
}

@inproceedings{albuquerque2024applying,
  title={Applying event classification to reveal the Estado da {\'I}ndia},
  author={Albuquerque, Gon{\c{c}}alo C and Souza, Marlo and Vieira, Renata and Ribeiro, Ana Sofia},
  booktitle={Proceedings of the 16th International Conference on Computational Processing of Portuguese-Vol. 1},
  pages={247--254},
  year={2024}
}

@inproceedings{sacramento2021joint,
  title={Joint event extraction with contextualized word embeddings for the portuguese language},
  author={Sacramento, Anderson da Silva Brito and Souza, Marlo},
  booktitle={Brazilian Conference on Intelligent Systems},
  pages={496--510},
  year={2021},
  organization={Springer}
}

@article{walker2006ace,
  title={Ace 2005 multilingual training corpus},
  author={Walker, Christopher and Strassel, Stephanie and Medero, Julie and Maeda, Kazuaki},
  journal={(No Title)},
  year={2006},
  publisher={Linguistic Data Consortium}
}

@article{abimannan2023ensemble,
  title={Ensemble multifeatured deep learning models and applications: A survey},
  author={Abimannan, Satheesh and El-Alfy, El-Sayed M and Chang, Yue-Shan and Hussain, Shahid and Shukla, Saurabh and Satheesh, Dhivyadharsini},
  journal={IEEE Access},
  volume={11},
  pages={107194--107217},
  year={2023},
  publisher={IEEE}
}

@article{aburomman2017survey,
  title={A survey of intrusion detection systems based on ensemble and hybrid classifiers},
  author={Aburomman, Abdulla Amin and Reaz, Mamun Bin Ibne},
  journal={Computers \& security},
  volume={65},
  pages={135--152},
  year={2017},
  publisher={Elsevier}
}

@article{masoudnia2014mixture,
  title={Mixture of experts: a literature survey},
  author={Masoudnia, Saeed and Ebrahimpour, Reza},
  journal={Artificial Intelligence Review},
  volume={42},
  number={2},
  pages={275--293},
  year={2014},
  publisher={Springer}
}

@inproceedings{kister-schirmer-2026-evaluating,
    title = "Evaluating Open-Source {LLM}s for Text Summarization and Named Entity Recognition in Long, Unstructured Text",
    author = "Kister, Pauline  and
      Schirmer, Miriam",
    editor = {Hamilton, Sil  and
      {\"O}hman, Emily  and
      Hicke, Rebecca M. M.  and
      Bizzoni, Yuri  and
      Bax, Axel  and
      Matthews, Jacob A.  and
      H{\"a}m{\"a}l{\"a}inen, Mika},
    booktitle = "Proceedings of the 6th International Conference on Natural Language Processing for the Digital Humanities",
    month = jul,
    year = "2026",
    address = "San Diego, USA",
    publisher = "Association for Computational Linguistics",
    url = "https://aclanthology.org/2026.nlp4dh-1.35/",
    doi = "10.18653/v1/2026.nlp4dh-1.35",
    pages = "390--410",
    ISBN = "979-8-89176-427-9"
}

@article{pepping2024reflections,
  title={Reflections on Encoding Languages in Historical Data: Working With the Multilingual Dimension of the Dutch East India Company Archives},
  author={Pepping, KW},
  journal={Journal of Open Humanities Data},
  volume={10},
  number={1},
  year={2024}
}

\clearpage 

\appendix

\begin{figure}[ht!]
    \includegraphics[width=17cm, height=10cm]{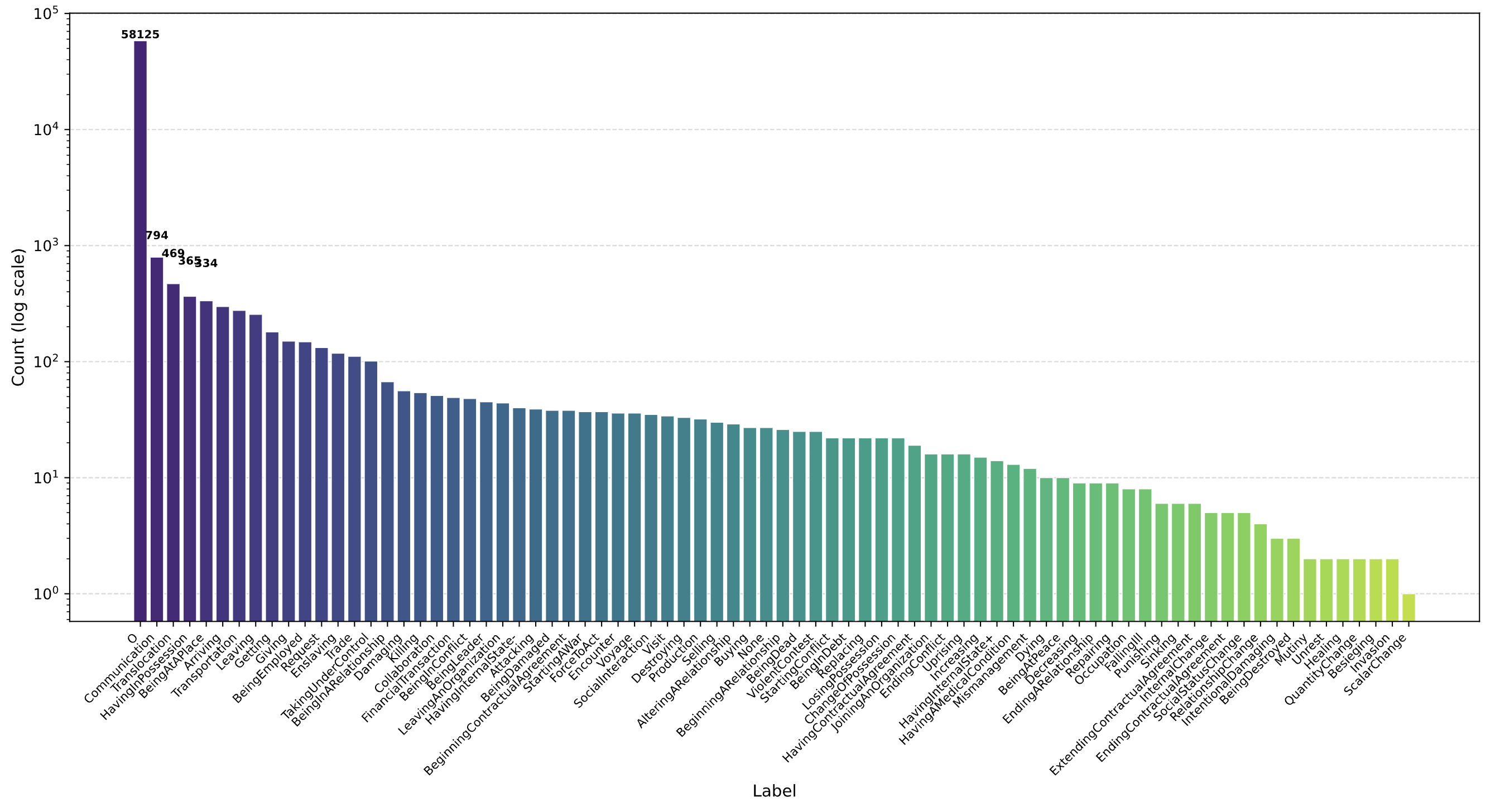}
    \caption{Label distribution in the expert annotated data of all classes (including 'O').}
    \label{fig:labeldistribution1}
\end{figure}

\begin{table}[]
    \centering
    \small
    \begin{tabular}{p{6cm} c c l c}
    \toprule
        Classes merged & Rep & Merge type & New name & New rep \\
    \midrule
        TakingUnderControl, Occupation & 101, 8 & Stat-Dyn & TakingUnderControl & 109 \\
        \addlinespace
        Attacking, Besieging, Invasion & 39, 2, 2 & Taxonomic & Attacking & 43 \\
        \addlinespace
        Destroying, BeingDestroyed & 33, 3 & Stat-Dyn & BeingDestroyed & 36 \\
        \addlinespace
        BeginningContractualAgreement, ExtendingContractualAgreement, EndingContractualAgreement, HavingContractualAgreement & 38, 6, 5, 19 & Stat-Dyn & HavingContractualAgreement & 68 \\
        \addlinespace
        Dying, BeingDead & 10, 25 & Stat-Dyn & BeingDead & 35 \\
        \addlinespace
        AlteringARelationship, EndingARelationship, RelationshipChange & 29, 9, 4 & Taxonomic & RelationshipChange & 42 \\
        \addlinespace
        StartingConflict, EndingConflict, BeingInConflict & 22, 16, 48 & Stat-Dyn & BeingInConflict & 86 \\
        \addlinespace
        Uprising, Mutiny, Unrest & 16, 2, 2 & Stat-Dyn & Unrest & 20 \\
        \addlinespace
        Repairing, Healing, HavingInternalState+ & 9, 2, 14 & Stat-Dyn & HavingInternalState+ & 25 \\
        \addlinespace
        FallingIll, HavingAMedicalCondition & 8, 13 & Stat-Dyn & HavingAMedicalCondition & 21 \\
    \bottomrule
    \end{tabular}
    \caption{Merged event classes. Rep = representation.}
    \label{tab:merged-classes}
\end{table}

\begin{table}[]
    \centering
    \small
    \begin{tabular}{p{15cm}}
    \toprule
    See the following example of an annotated sentence: {example}. Create a sentence in Early Modern Dutch in which {label} is transported. This is the definition of {label}: {definition}. Write a sentence that could be found in the Archives of the Dutch East India Company (1602-1799). Make sure to adjust your style to the Dutch East India Company. Put the word that refers to the transportation in square brackets like in the example. \\ \hdashline
    See the following example of an annotated sentence: {example}. Create a sentence in Early Modern Dutch in which {label} is transported. This is the definition of {label}: {definition}. Write a sentence that could be found in the Archives of the Dutch East India Company (1602-1799). Make sure to adjust your style to the Dutch East India Company. Put the word that refers to the transportation in square brackets like in the example. Do not repeat the definition as given in the prompt, only use it for your own context. \\ \hdashline
    See the following example of an annotated sentence: {example} Create a sentence in Early Modern Dutch in which a translocation is taking place with a {label}. This is the definition of {label}: {definition}. Write a sentence that could be found in the Archives of the Dutch East India Company (1602-1799). Make sure to adjust your style to the Dutch East India Company. Put the word that refers to the translocation in square brackets like in the example. Make sure to vary in syntax.\\ \hdashline
    See the following example of an annotated sentence: {example} Create a sentence in Early Modern Dutch in which a translocation is taking place with a {label}. This is the definition of {label}: {definition}. Write a sentence that could be found in the Archives of the Dutch East India Company (1602-1799). Make sure to adjust your style to the Dutch East India Company. Put the word that refers to the translocation in square brackets like in the example. A translocatiion could for example be an arrival, a departure, a voyage or a transport. Make sure to vary in syntax.\\ \hdashline
    Create a sentence in Early Modern Dutch in which a translocation is taking place with a {label}. This is the definition of {label}: {definition}. Write a sentence that could be found in the Archives of the Dutch East India Company (1602-1799). Make sure to adjust your style to the Dutch East India Company. Put the word that refers to the translocation in square brackets. Do not start your sentence with a time indication.\\ \hdashline
    See the following example of an annotated sentence: {example}. Create a similar sentence, that differs in syntax and vocabulary, in Early Modern Dutch in which a translocation is taking place. A translocation could for example be an arrival, a departure, a voyage or a transport. Write a sentence that could be found in the Archives of the Dutch East India Company (1602-1799). Make sure to adjust your style to the Dutch East India Company. Put the word that refers to the translocation in square brackets like in the example. \\ \hdashline
    Create a sentence in Early Modern Dutch in which {name} is travelling to or from {location}. The way {name} is travelling can be inspired by their profession, which is {function}. Write a sentence that could be found in the Archives of the Dutch East India Company (1602-1799). Make sure to adjust your style to the Dutch East India Company. Put the word that refers to the translocation in square brackets.\\ \hdashline
    Create a sentence in Early Modern Dutch in which {name} is travelling to or from {location}. The way {name} is travelling can be inspired by their profession, which is {function}. Write a sentence that could be found in the Archives of the Dutch East India Company (1602-1799). Make sure to adjust your style to the Dutch East India Company. Put the word that refers to the translocation in square brackets. Do not start the sentence with {name}.\\ 
    \bottomrule
    \end{tabular}
    \caption{All prompts used to create synthetic data}
    \label{tab:prompts}
\end{table}

\begin{table}[]
    \centering
    \small
    \begin{tabular}{p{15cm}}
    \toprule
    wordende door de onderkooplieden Jevers en van Bijland aan de vrlreep gerecibieert, en bij de hand naar den Heer Commissaris [geleid] bestaande in Aria soeradikarsa, Aria wangsanagara Aria ong Jaija, en den Ingabeij Ratoe Bagoes Pringalaija\\ \hdashline
    Den 27=en Januarij 1625 sijn dan hier in sloote naer t vaderslandt [vertrocken], ondert Commandement, vanden E: Cornelis Geuertss de Scheepen hollandia Goudaende Middelburch. \\ \hdashline
    Vant vaderslandt sijn hier Godtloff alle de Schepen uijt gesondert alckmaer int Gexel behouden [aengecomen] Namentlijck . T Jacht de Hase den 4 Junij 1625 T schip de Cameel den 19: septemb = r \\ \hdashline
    Is bij onsen scheeps Raet goet gevonden om [voort te seijlen] naer de soute Eijlande om hem aldaer te vinden ofte een dach off drie te verwachten \\ \hdashline
    deen 25 . november sijn wij [gepasseert] het Eijlant palma ende vernomen dat ons Esels hooft gans veermolsent ( was soo dat wij L. tgroote marsseyjl niet dorsten voeren den 10 december sin wij [gearrineert] opde rede voor Jlijo de maijo alwaer wij een ander Esels hooft hebben toe gestelt\\ \hdashline
    man die binnen boord bleeven , in de boot aan land [gevlugt] , en daar soolange verbleven waren tot tijd enwijle sagen dat hetselve bij hare chialoup — aan boord gelegen , daar eenige uuren aandoende geweest dog die nae dato weder [verlaten] , en ['t zee gat gekosen] hadde , als wanneer hun ook weder naar derwaarts [begeven] , en aan boord komende door de twee aldaar verblevene maats te verstaan gegeven was \\ \hdashline
    t beste benevens diverse provi sien , Equipagie goederen \& t „ a tesamen omtrend ter waardije van 400 : thailen daar [uijtgeligt] envoor een goede buijt [mede gevoert] hadden \\ \hdashline
    t beste benevens diverse provi sien , Equipagie goederen \& t „ a tesamen omtrend ter waardije van 400 : thailen daar [uijtgeligt] envoor een goede buijt [mede gevoert] hadden waar nae demallabaren met d „ o chialoup ook [terug gekeert] , en weder op de atchinse rheede [gelopen]\\ 
    \bottomrule
    \end{tabular}
    \caption{All example annotations used to fill the prompt templates' example slots}
    \label{tab:examples}
\end{table}

\begin{table}[ht]
\centering
\begin{tabular}{lccc:ccc:c}
\toprule
 & \multicolumn{3}{c}{ROBE with synth} & \multicolumn{3}{c}{ROBE+ with synth} & \\
\cmidrule(lr){2-4} \cmidrule(lr){5-7}
 & P & R & F1 & P & R & F1 & Support \\
\midrule
overall & 0.45 & 0.18 & 0.26 & 0.30 & 0.24 & 0.27 & 626 \\
translocation & 0.77 & 0.45 & 0.57 & 0.65 & 0.63 & 0.64 & 208 \\
violence & 0.38 & 0.08 & 0.13 & 0.32 & 0.12 & 0.18 & 73 \\
\bottomrule
\end{tabular}
\caption{Results across event classes for ROBE with a classifier for translocation classes trained on synthetically augmented data and ROBE+ with the same classifier for translocation classes}
\label{tab:me_combined_synth}
\end{table}

\begin{table}[ht]
\centering
\begin{tabular}{lccc:ccc:ccc:ccc:c}
\toprule
 & \multicolumn{6}{l}{\textbf{Zero-shot}} & \multicolumn{7}{l}{\textbf{Few-shot}} \\
\midrule
 & P & R & F1 & P & R & F1 & P & R & F1 & P & R & F1 & Support \\
\midrule
overall & 0.07 & 0.04 & 0.05 & 0.07 & 0.05 & 0.06 & 0.03 & 0.04 & 0.04 & 0.03 & 0.04 & 0.04 & 626 \\
translocation & 0.21 & 0.12 & 0.15 & 0.20 & 0.11 & 0.14 & 0.14 & 0.13 & 0.13 & 0.15 & 0.13 & 0.14 & 208 \\
violence & 0.42 & 0.11 & 0.17 & 0.42 & 0.07 & 0.12 & 0.14 & 0.08 & 0.10 & 0.18 & 0.05 & 0.08 & 73 \\
\bottomrule
\end{tabular}
\caption{Macro results across event classes for two runs of gpt-5.1 in zero-shot setting and two runs in few-shot setting with temperature=0. Metric scores are calculated in the same way as in Table \ref{tab:results}.}
\label{tab:gpt-results}
\end{table}

\end{document}